\PassOptionsToPackage{table,xcdraw}{xcolor}
\documentclass[sigconf]{acmart}
\renewcommand\footnotetextcopyrightpermission[1]{} 
\usepackage{array, makecell, placeins}
\usepackage[T1]{fontenc}
\usepackage{lmodern}
\usepackage{float}

\setcopyright{none}
\begin{document}

\author{Muhammad Hisyam Zayd}
\affiliation{
    \institution{Teknik Informatika, Fakultas Ilmu Komputer, Universitas Brawijaya}
    \city{Malang}
    \country{Indonesia}}
\email{hisyamzayd@gmail.com}

\author{Novanto Yudistira}
\affiliation{
    \institution{Teknik Informatika, Fakultas Ilmu Komputer, Universitas Brawijaya}
    \city{Malang}
    \country{Indonesia}}
\email{yudistira@ub.ac.id}

\author{Randy Cahya Wihandika}
\affiliation{
    \institution{Teknik Informatika, Fakultas Ilmu Komputer, Universitas Brawijaya}
    \city{Malang}
    \country{Indonesia}}
\email{rendicahya@ub.ac.id}

\title{Image Colorization using U-Net with Skip Connections and Fusion Layer on Landscape Images}

\begin{abstract}
    We present a novel technique to automatically colorize grayscale images that combine the U-Net model and Fusion Layer features. This approach allows the model to learn the colorization of images from pre-trained U-Net. Moreover, the Fusion layer is applied to merge local information results dependent on small image patches with global priors of an entire image on each class, forming visually more compelling colorization results. Finally, we validate our approach with a user study evaluation and compare it against state-of-the-art, resulting in improvements.
\end{abstract}


\begin{teaserfigure}
    \includegraphics[width=\textwidth]{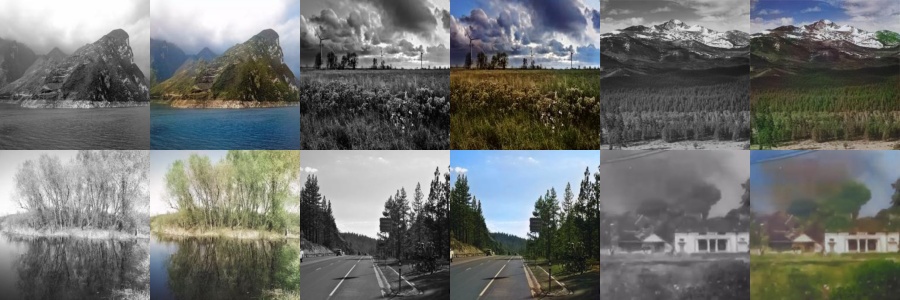}
    \caption{
        Colorization result from black and white landscape images and photographs by our model. \textit{Islet landscape image} (left top), \textit{Wind Farm landscape image} (middle top), \textit{Rocky Mountain National Park, Colorado, ca. in 1941-1942 by Ansel Adams} (right top), \textit{Marsh landscape image} (left bottom), \textit{Forest road landscape image} (middle bottom), \textit{Real footage of Operation Product of Dutch Military in Malang, Indonesia in 1947} (right bottom)
    }
    \Description{colorization result - teaser}
    \label{fig:teaser}
\end{teaserfigure}

\maketitle
\pagestyle{plain}

\section{Introduction and Related Work}
    Automatic image colorization using deep learning has become a popular research object since 2016. Of the popular image colorization studies, we summarize three shortcomings and suggestions for improvement of the model. \textbf{First} is the need to use a neural network model specifically fine-tuned from classification encoder so that the colorization results on each object are expected to be more consistent. For example, the color of the grass commonly has to be green or perhaps yellow but never be colored by blue \cite{hwang2016image}. \textbf{Second} is the need to always use a neural network with classification + regression or \textbf{proposed model} for colorization. Recent studies have shown that the proposed model outperforms the \textbf{baseline or regression model only} by creating a stronger intensity of the color for each object \cite{iizuka2017globally} \cite{hwang2016image}. \textbf{The third} is the need to use only one type of image or dataset for each model, such as landscape images or human's created object images. Colorization with deep learning is data-driven and often creates inconsistent color results for different types of images \cite{iizuka2017globally}. One image type that shows resulting more stable, compelling, and reasonable color compared to other types even the dataset is consists of various types of images is \textbf{landscape images} \cite{hwang2016image} \cite{baldassarre2017deep} \cite{appelgrenevaluation}.
    \\~\\
    Our work is similar to \cite{IizukaSIGGRAPH2016} with the one of differences is that we try to propose a new model based on these evaluations \textbf{by combining U-Net with the Fusion Layer}. U-Net is the most popular Convolutional Neural Network model that works explicitly for image segmentation \cite{ronneberger2015u}. Recent studies show that even the simplified/downscaled U-Net can be used as a colorization model, resulting in enough compelling images compared to the state of the art \cite{appelgrenevaluation}. Meanwhile, the Fusion layer is a classification model type of neural network introduced by \cite{iizuka2017globally} for automatic colorization. The fusion Layer + regression model also shows better performance than the baseline model by creating more stable and compelling colors.
    \\~\\
    On the other hand, there is another type of classification model for automatic colorization namely \textbf{binning function} by \cite{zhang2016colorful} and \cite{hwang2016image}. The difference is \cite{zhang2016colorful} divide color space into 313 different buckets, while \cite{hwang2016image} divides color space into 50 different buckets. Based on \cite{appelgrenevaluation}, the binning function of 313 different buckets with simplified U-Net \textbf{took up an excess of 200 GB in memory} when only using 64x64 size of an image. Meanwhile, the binning function of 50 different buckets is also said to be memory-intensive \cite{hwang2016image}. Therefore we tried to utilize Fusion Layer as our classification model for automatic colorization as an alternative for memory space excuse and proved that Fusion Layer with U-Net and Resnet-34 as an encoder layer of U-Net \textbf{only took memory space around 60 GB, three times more efficient} than binning function with simplified U-Net.
    \\~\\
    Our study found that the best configuration of our model is U-Net with \textit{pre-trained} Resnet34 as an encoder layer of U-Net and Fusion Layer + regression model (proposed model) with batch number is 64, using Adam as an optimization function with \textit{lr} value is 0.01, and total epoch of training is 10. We try many configurations by combining the proposed model and baseline model with a variation of batch number, optimization function type, and \textit{lr} value. We used 682.382 images for training, 13700 images as data for validation, and 1370 images for the test, consisting of 137 total classes. We subjectively selected the class by considering whether it is related to the 'landscape' image. Raw dataset is taken from Place365 \cite{khosla} with 365 classes in total with 1,817,700 data training, and 3,650 images as data testing.
    \\~\\
    We try to evaluate our best model by comparing it against the state-of-the-art. The evaluation consisted of comparing \textit{user study evaluation} 's accuracy value from each model, comparing the average Mean Absolute Error (MAE) value between colorized images and real/ground truth images from each model, and comparing the colorization result of real past black and white images from each model. The result is that our model can compete with state-of-the-art by resulting in \textbf{the lowest average MAE value, the second-highest accuracy result value of \textit{user study evaluation}, and the most compelling color on real past black and white images (figure \ref{fig:teaser}).}

\section{How Automatic Colorization Works}
    Based on recent studies, colorization with deep learning works by creating colorized images from black and white images as input using deep neural networks. As deep learning is ’learning’ how well the output is, colorized images must be compared with real colored images. Therefore, we need two types of color from one image, black and white, as input and colored image as a ground truth image for final evaluation.
    \\~\\
    \textbf{CIE La*b*} is one type of color space that consists of L for lightness perceptual of an image representing the intensity of monochrome color and a*b* for chrominance perceptual of an image representing the red, green, blue, and yellow color. On-screen output, L channel’s value ranged between 0 and 100, and a*b* channel’s value ranged between -128 and 127 \cite{wikipedia_cielab}. L channel is used as an input for deep learning, while a* and b* channel used to compare deep learning is a* and b* output channel. Previous studies have shown that CIE La*b* color space performs better than any other color space type, resulting in a more reasonable colorized image. Therefore we choose CIE La*b* as color space for our colorization model.
    \\~\\
    L channel and a* b* channel of an image have to be normalized before entering the deep learning. This is because deep learning can result in value in any real numbers, while a* and b* output channels have to be ranged between -128 and 127. \textbf{That is why we need to normalize the L and a* b* channel as ranged between 0 - 1} \cite{sola1997importance}. After propagation, a* b* output channel value has to be back to 0 - 1 range with \textbf{sigmoid function}. The sigmoid function will return any value between 0 and 1 with any real number as input \cite{han1995influence}. a* and b* output channel with L channel (input) of CIE La*b* color space is converted back to the RGB channel to create a new colorized image.

    \begin{figure*}[h!]
        \centering
        \includegraphics[scale=0.3]{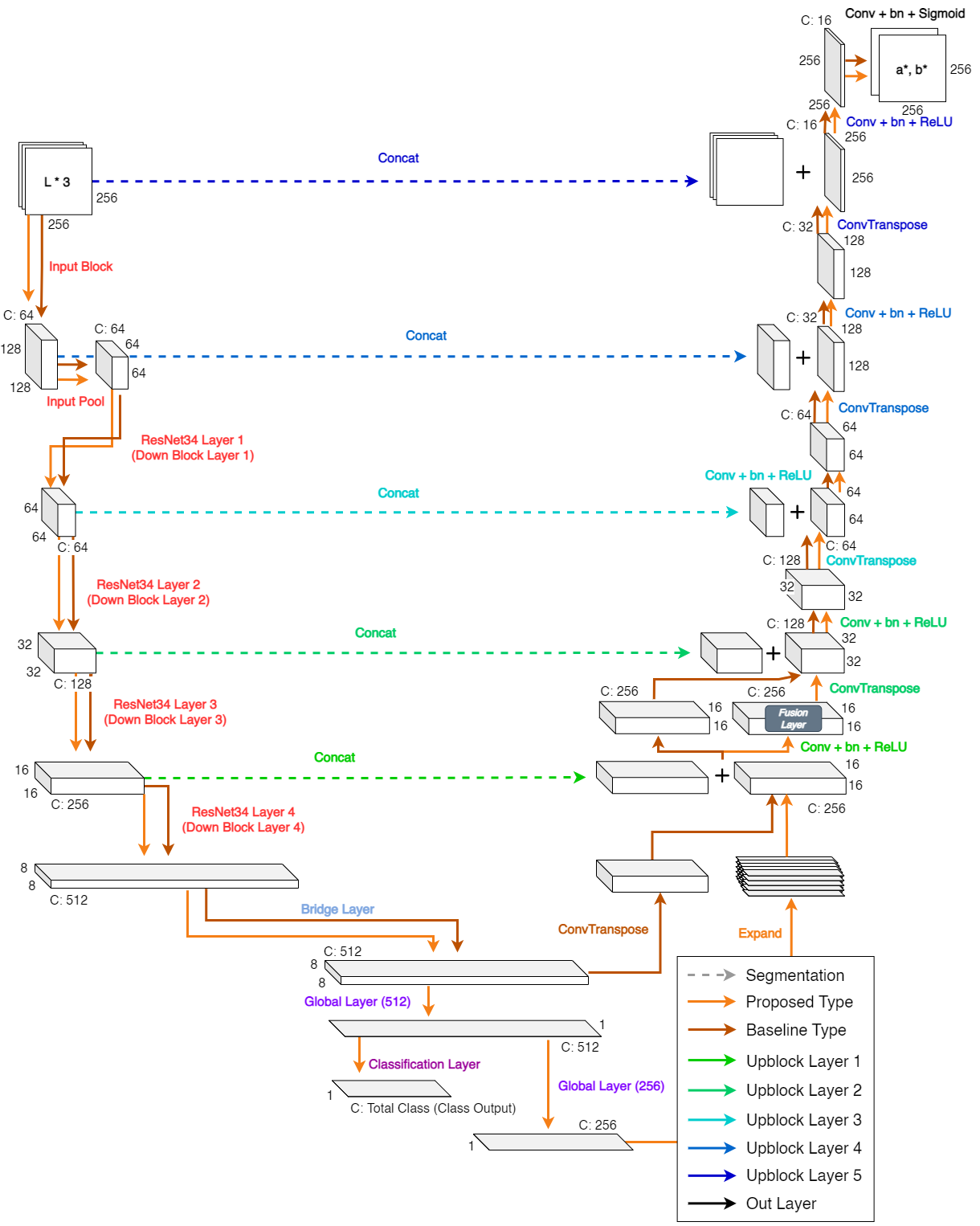}
        \caption{
            Overview of U-Net model with Fusion Layer and Resnet34 as encoder part
        }
        \label{fig:model_overview}
        \Description{Overview of Model}
    \end{figure*}

 \begin{table*}[h!]
        \caption{
            Architecture Design of U-Net model with Fusion Layer and Resnet34 as encoder part (K, O, I, W, H are kernel size, output channel, input channel, output width, and output height, respectively)
        }
        \label{tab:arch_design}
        \resizebox{\textwidth}{!}{
            \begin{tabular}{
                c|c|c|c|c|c|c
            }
                \textbf{Network Path} & \textbf{Layer} & \textbf{Size of \textit{K x K x O x I}} & \textbf{\makecell{Output Size \\ \textit{(O x W x H)}}} & \textbf{Batch Norm.} & \textbf{Activation} & \textbf{Parameters} \\
            \hline
                & Input Block & 7 x 7 x 64 x 3 & 64 x 128 x 128 & True & ReLU & 9,536 \\
                & Input Pool & 3 x 3 x 64 x 64 & 64 x 64 x 64 & False & - & - \\
                \textbf{Encoder} & Down Block Layer 1 & 3 x 3 x 64 x 64 & 64 x 64 x 64 & True & ReLU & 221,952 \\
                & Down Block Layer 2 & 3 x 3 x 128 x 64 & 128 x 32 x 32 & True & ReLU & 1,116,416 \\
                & Down Block Layer 3 & 3 x 3 x 256 x 128 & 256 x 16 x 16 & True & ReLU & 6,822,400 \\
                & Down Block Layer 4 & 3 x 3 x 512 x 256 & 512 x 8 x 8 & True & ReLU & 13,114,368 \\
            \hline
                & Bridge Layer & 3 x 3 x 512 x 512 & 512 x 8 x 8 & True & ReLU & 4,721,664 \\
                \textbf{Middle Part} & Global Layer & \makecell{1 x 1 x 512 x 512, \\ 1 x 1 x 256 x 512} & 512 x 1, 256 x 1 & True & ReLU & 34,215,168 \\
                & Classification Layer & 1 x 1 x Total Class x 512 & Total Class x 1 & True & Softmax & 167,323 \\
            \hline
                & Up Block Layer 1 & - (Fusion Layer) & 256 x 16 x 16 & True & ReLU & 1,771,008 \\
                & Up Block Layer 2 & 2 x 2 x 128 x 256 & 128 x 32 x 32 & True & ReLU & 574,336 \\
                \textbf{Decoder} & Up Block Layer 3 & 2 x 2 x 64 x 128 & 64 x 64 x 64 & True & ReLU & 143,808 \\
                & Up Block Layer 4 & 2 x 2 x 32 x 64 & 32 x 128 x 128 & True & ReLU & 45,280 \\
                & Up Block Layer 5 & 2 x 2 x 16 x 32 & 16 x 256 x 256 & True & ReLU & 7,200 \\
                & Conv Out Layer & 1 x 1 x 2 x 16 & 2 x 256 x 256 & True & Sigmoid & 38 \\
            \hline
                \multicolumn{6}{c|}{\textbf{Total Parameters}}&\textbf{62,930,497} \\
            \hline
            \end{tabular}
        }
    \end{table*}

\section{Our Model - U-Net with Fusion Layer}
    U-Net is a deep neural network with an autoencoder type. Autoencoder combines an encoder function that converts the input data into a different representation and a decoder function that converts the new representation back into the original format. This approach allows the model to make new representations based on many preserved information \cite{goodfellow2016deep}. The encoder part consists of many convolutional neural networks that increase the scale of the input and the channel of the input. While decoder part consists of many convolutional neural networks and convolutional transpose neural networks to return the scale and channel size of the encoder output to its original size. In U-Net, the encoder and decoder part is connected with \textbf{a bridge layer} that consists of 2 times 2d convolution layer. U-Net also consists of \textbf{stacking method}, combining results from every encoder part into every decoder part matching with the sequence.
    
    \begin{equation}
    \label{eqn:eq1}
        \textbf{Y}^{fusion}_{u, v} = ReLU \left( b + W \begin{bmatrix} \textbf{Y}^{expanded\_global}\\\textbf{Y}^{last\_encoder}_{u, v}\end{bmatrix} \right)
    \end{equation}
    
    As in Iizuka et al., \textbf{Fusion Layer} (eq. \ref{eqn:eq1}) combines both global features network and mid-level features network result into one single dimension object. Both global features network and mid-level features network are implementations of encoder part of ColorNet. We try to add this Fusion Layer in the first decoder part of our U-Net, combining the result of the last encoder part with the result of \textbf{the \textit{expanded} global layer result}. We add \textbf{the global layer} and \textbf{classification layer} after the bridge layer and before the first decoder part as to how the classification model in Iizuka et al. works.

    This transformation towards U-Net by extending its existing loss function or adding a new computational model such as classification layer also has been done by previous studies and proved to create more great results than plain/classical U-Net does \cite{hakim2019u} \cite{yudistira2020prediction}. Illustration of where Fusion Layer, global feature layer, and classification layer located and the sequence of works is shown in \textbf{figure ~\ref{fig:model_overview}}.

    We adapt the transfer learning method to our model. The goal of transfer learning is to improve learning in the target task by leveraging knowledge from the source task \cite{torrey2010transfer}. In the CNN model, the weight of each layer is the knowledge of the model. Transfer learning can be done by replacing the encoder part of U-Net with another model by doing adjustment \cite{pravitasari2020unet}. We use \textbf{the \textit{pre-trained} Resnet34 model} as our encoder part of U-Net. We cannot use more than 34 layer types of Resnet because it is exceeding the memory limit of our computational resources, and the performance of 34 layers is better than 18 layers as mentioned in \cite{he2016deep}. Before entering the encoder part, we add \textbf{input block} and \textbf{input pool} which is \textit{conv1} and max-pool layer from \textit{conv2\_x} of Resnet34. The total number of parameters of proposed model is shown in table \ref{tab:arch_design}.

    Standard U-Net uses two times 2d convolutional neural network for each encoder part, followed by max-pooling. Our U-Net model uses 2d convolutional nets \textbf{with increased strides} for reducing the size of the feature maps instead of using max-pool layers. This can be increasing the spatial support of each layer \cite{iizuka2017globally}. 
 
    Another modification for our U-Net model is in every 2d convolutional nets are always followed by batch normalization and rectified linear unit or \textit{ReLU} as an activation function. Using batch normalization will dramatically reduce training convergence time and improve accuracy \cite{ioffe2015batch} as also does \textit{ReLU} will significantly accelerate training convergence time \cite{krizhevsky2012imagenet}.

    Based on recent studies, regression error calculation commonly use the \textbf{Mean Square Error (MSE)} function while classification error calculation commonly uses \textbf{the Cross-Entropy Loss} function \cite{iizuka2017globally} \cite{hwang2016image} \cite{baldassarre2017deep} \cite{appelgrenevaluation}. In our proposed methods, there are two types of model: \textbf{Proposed w/o Fusion layer} and \textbf{Proposed model}. \textbf{Proposed} means the training model uses Fusion layer with loss function of  MSE generated from last decoder layer and classification loss of Softmax Cross-Entropy generated from \textbf{classification layer}. While \textbf{Proposed w/o Fusion layer} only consists of MSE loss generated from last decoder layer. The MSE is also believed to be more meaningful than the commonly used criterion, the residual sum of squares \cite{allen1971mean} and the Cross-Entropy Loss has shown significant and practical advantages over squared-error function \cite{kline2005revisiting}.

\begin{figure*}[h!]
        \centering
        \includegraphics[width=\linewidth]{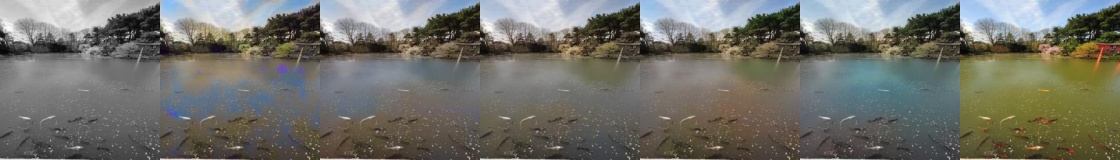}
        \caption{
            Comparison of Colorization Result by All Configurations. From Left to Right: \textit{Input, Proposed 1, Proposed 2, Proposed 3, Proposed 4, Proposed 2 w/o Fusion layer} and \textit{Ground Truth}
        }
        \label{fig:config_comparison}
        \Description{configurations compare result}
    \end{figure*}
    
     \begin{figure*}[h!]
        \centering
        \includegraphics[scale=0.5]{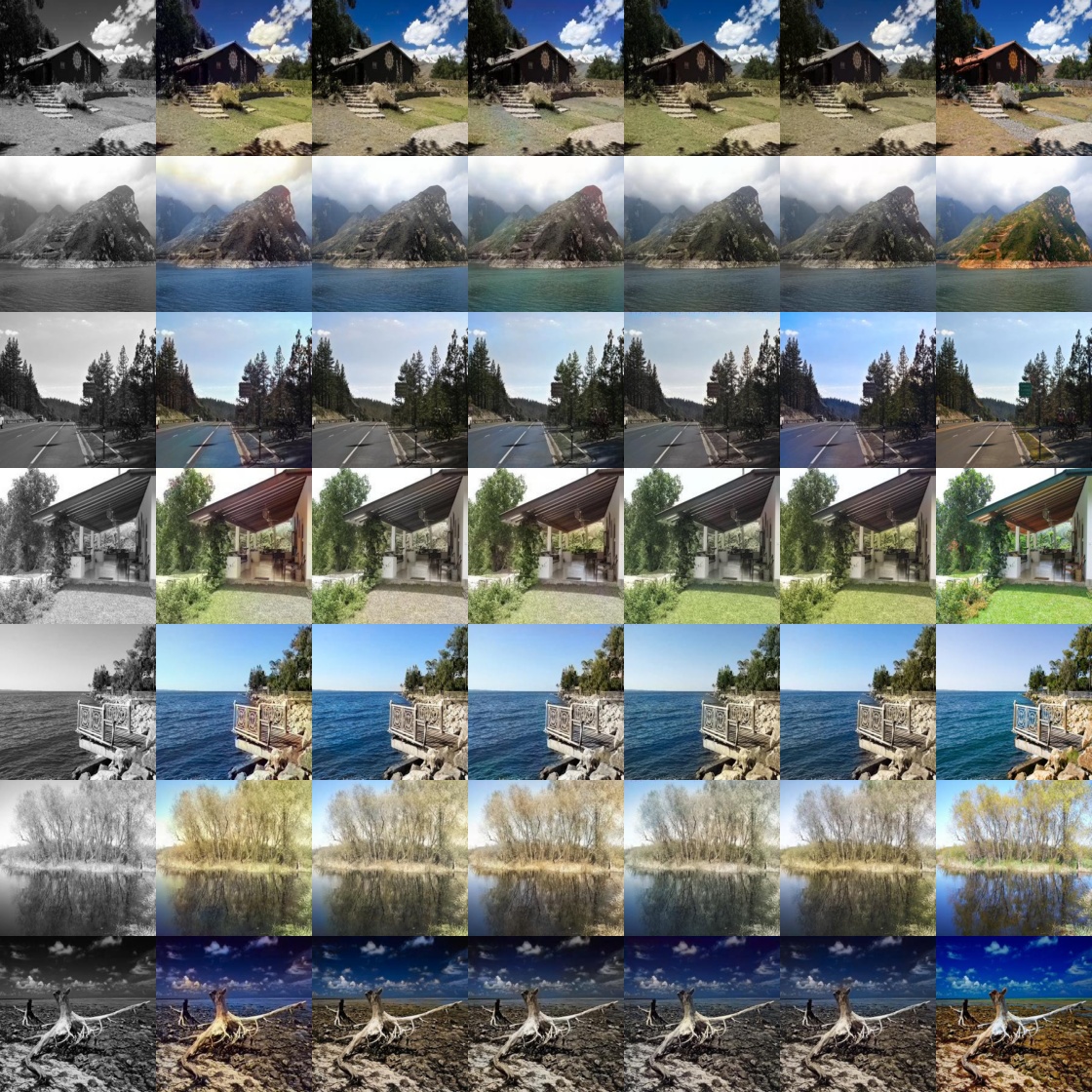}
        \caption{
            Best 7 Colorization Result, Subjectively Selected. From Left to Right: \textit{Input, Zhang (2016) et al., Zhang (2017) et al., ColorNet, Proposed 2 w/o Fusion layer, Proposed 3,} and \textit{Ground Truth}
        }
        \label{fig:best_7_result}
        \Description{Best 7 Result}
    \end{figure*}

\section{Experimental Results and Discussion}
    \subsection{Variation of Model Configuration}
        First, we try to find the best combination of variation of hyperparameters, including \textbf{batch number, optimization function, learning rate value, and model type (proposed or proposed w/o Fusion layer)}. Then, each model configuration will be trained as ten total epochs, require about 40 hours of running time, and be tested with 1370 images. All configurations are listed below:
        
        \begin{itemize}
            \item \textbf{Proposed 1}: use Adadelta optimization function with \textit{lr} value of 0.03 and batch number of 64
            \item \textbf{Proposed 2}: use Adam optimization function with \textit{lr} value of 0.01 and batch number of 16
            \item \textbf{Proposed 3}: use Adam optimization function with \textit{lr} value of 0.01, batch number of 16, and divide loss/error value of classification layer by 100 as ColorNet \cite{iizuka2017globally}
            \item \textbf{Proposed 4}: use Adam optimization function with \textit{lr} value of 0.01 and batch number of 64
            \item \textbf{Proposed 2 w/o Fusion layer)}: use Adam optimization function with \textit{lr} value of 0.01 and batch number of 16
        \end{itemize}
        
        Experiment is conducted by comparing MAE on various aforementioned configurations as listed in \textbf{table \ref{tab:config_test_result}}. The best configuration is colorized as shown as \textbf{a green mark} and the second-best MAE is colorized in \textbf{a yellow mark}. 
        
        \begin{table}[h]
            \centering
            \caption{Configuration Test Result}
            \resizebox{\columnwidth}{!}{
                \begin{tabular}{c c c c}
                     \textbf{Config. Name} & \textbf{Avg. MAE} & \textbf{Avg. MSE a*b*} & \textbf{Avg. MSE La*b*} \\
                    \hline
                    \textbf{Proposed 1} & 0.054470 & 0.003454 & 0.008081 \\
                     \textbf{Proposed 2} &  0.044688 &  0.002233 &  0.005295 \\
                     \cellcolor[HTML]{fdd835}
                     \textbf{Proposed 3} &  \cellcolor[HTML]{fdd835}0.043767 &  \cellcolor[HTML]{fdd835} 0.002177 & \cellcolor[HTML]{fdd835} 0.005151 \\
                    
                    \textbf{Proposed 4} & 0.045120 & 0.002367 & 0.005509 \\
                    
                    \cellcolor[HTML]{64dd17} \textbf{Proposed 2 w/o Fusin layer} & \cellcolor[HTML]{64dd17} 0.040888 & \cellcolor[HTML]{64dd17} 0.001949 & \cellcolor[HTML]{64dd17} 0.004554 \\
                    \hline
                \end{tabular}
            } 
            \label{tab:config_test_result}
        \end{table}
        
        Based on the experiment result on configurations above (see \textbf{figure ~\ref{fig:config_comparison}}), Proposed 2 w/o Fusion layer produces lowest MAE, however, \textbf{Proposed 3 creates the most stable and closest color result to ground truth compared to another proposed model}. Even though Proposed 2 w/o Fusion layer has the lowest error rate value, the colorization results are still far from ground truth such as the color of lake is rather blue than green as shown in \textbf{figure \ref{fig:config_comparison}} and blue sky color by Proposed 3 is stronger than Proposed 2 w/o Fusion layer in \textbf{figure \ref{fig:best_7_result}}. After all, U-Net with Fusion layer's color result is more stable corresponding to the ground-truth and colorful.

\begin{figure*}[h!]
        \centering
        \includegraphics[scale=0.41]{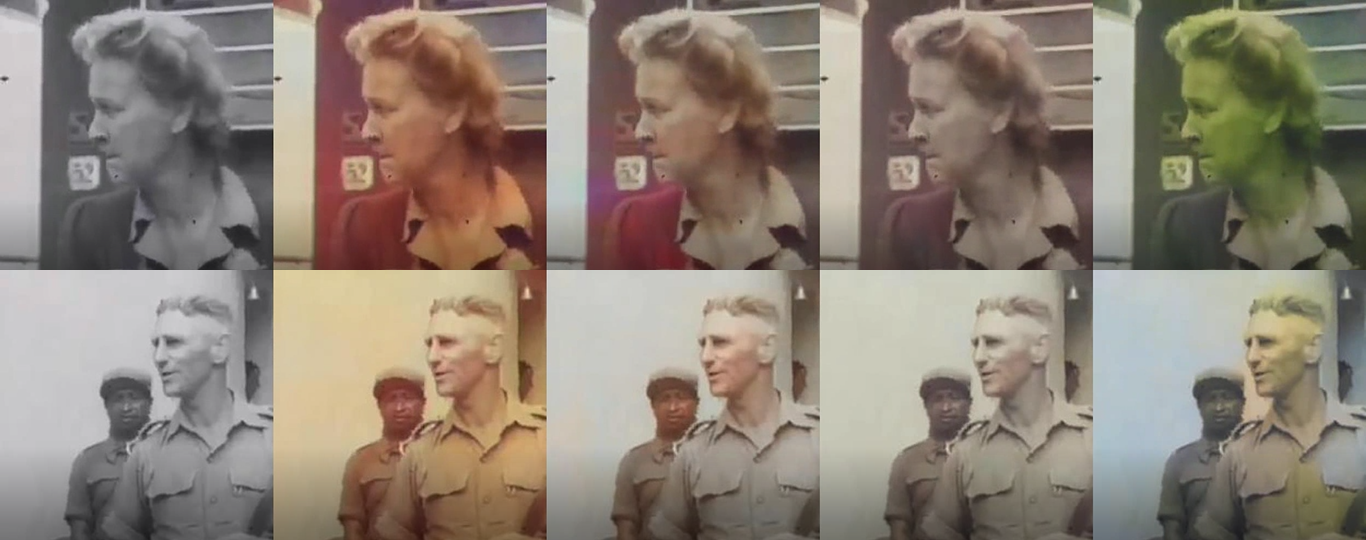}
        \caption{
            Colorization Result on Human Object of Real Past Footage. From Left to Right: \textit{Input, Zhang (2016) et al., Zhang (2017) et al., ColorNet,} and \textit{Proposed 3}
        }
        \label{fig:past_photo_human_obj}
        \Description{Past Photo of Human Obj}
    \end{figure*}
    
     \begin{figure*}[h!]
        \centering
        \includegraphics[scale=0.41]{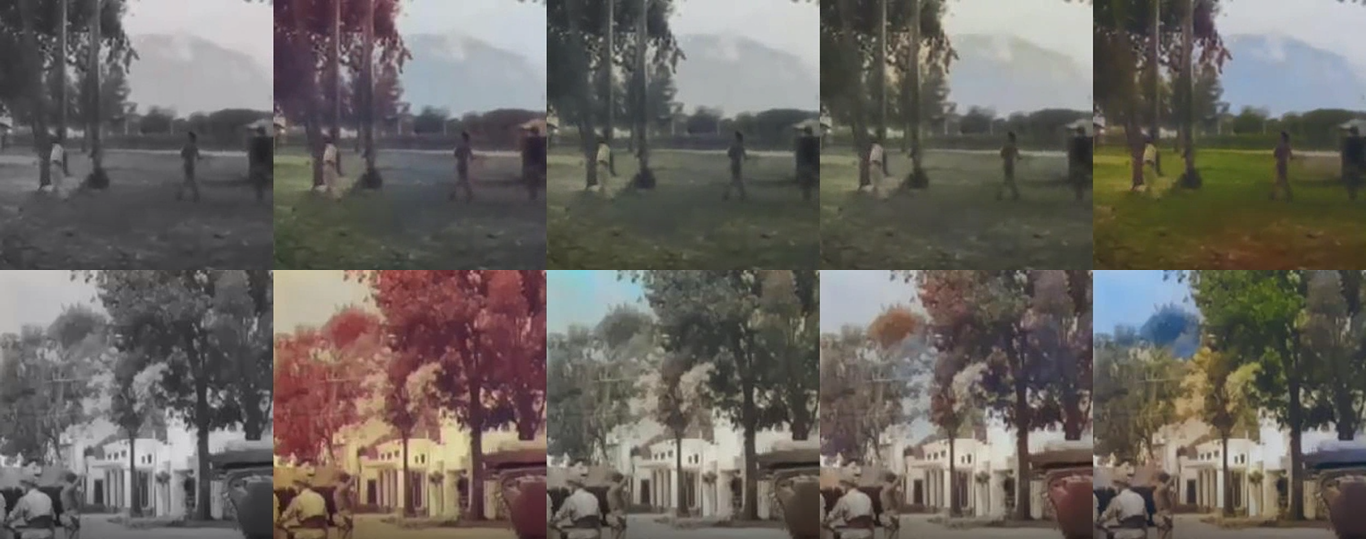}
        \caption{
            Best Colorization Result of Real Past Footage. From Left to Right: \textit{Input, Zhang (2016) et al., Zhang (2017) et al., ColorNet,} and \textit{Proposed 3}
        }
        \label{fig:past_photo_best}
        \Description{Best Past Photo}
    \end{figure*}

    \subsection{Comparing Against the State of the Art}
        \textit{User study evaluation} is a technique proposed by Iizuka et al. to evaluate and compare colorization results from models. Iizuka et al. provide 500 images in total from each type: ground-truth image, baseline model colorization result, and proposed model colorization result. Respondents have to decide for each 500 image whether \textbf{real ground truth images or fake} resulting from the colorization model without knowing the exact colorization model for each image. In our experiment, \textbf{we use 100 total images} for each type: Zhang (2016) et al. colorization result \cite{zhang2016colorful}, Zhang (2017) et al. colorization result \cite{zhang2017real}, and Proposed 4 colorization results are included. \textbf{We can not use ColorNet} \cite{iizuka2017globally} as an additional model for comparison because its pre-trained model is not provided at that time experiment was conducted. The total of our respondents is 17. Based on average accuracy’s result (see \textbf{table \ref{tab:avg_user_study_result}}), \textbf{Proposed 4 outperforms the Zhang (2016) et al. result and is comparable to Zhang (2017) et al. colorization model}.
        
        \begin{table}[h]
            \caption{User Study Evaluation Result}
            \label{tab:avg_user_study_result}
            \begin{tabular}{c c}
                \textbf{Name of Model} & Average Accuracy \\
                \hline
                    \textbf{Zhang (2016) et al.} & 44.5294 \% \\
                    \cellcolor[HTML]{64dd17} \textbf{Zhang (2017) et al.} & 56.8235 \% \\
                    \cellcolor[HTML]{fdd835} \textbf{Proposed 4} & 48.1765 \% \\
                \hline
            \end{tabular}
        \end{table}
        
        Next we conduct comparison of the average of MAE throughout the test dataset of each model to confirm the difference between colorization result and ground-truth image. MAE is a more natural measure of average error, unlike RMSE (Root Mean Square Error) for average-model-performance error \cite{willmott2005advantages}. The evaluation result (see \textbf{table \ref{tab:avg_mae_result}}) shows that our best model are Proposed 3 and Proposed 2 w/o Fusion layer (\textbf{still has the lowest average MAE value compared to previous studies}). We choose top 7 image colorization results by all models and shown in \textbf{figure \ref{fig:best_7_result}}
        
        \begin{table}
            \caption{MAE Test Result}
            \label{tab:avg_mae_result}
            \begin{tabular}{c c}
                \textbf{Name of Model} & Average MAE \\
                \hline
                    \textbf{Zhang (2016) et al.} & 0.05311835 \\
                    \textbf{Zhang (2017) et al.} & 0.04648737 \\
                    \textbf{ColorNet} & 0.04631426 \\
                    \cellcolor[HTML]{fdd835} \textbf{Proposed 3} & 0.04376666 \\
                    \cellcolor[HTML]{64dd17} \textbf{Proposed 2 w/o Fusion layer} & 0.04088821 \\
                \hline
            \end{tabular}
        \end{table}

        The last evaluation compares the colorization result of real past black and white images from each model. We found an old video titled "Operation Product of Dutch Military at 1947 in Malang, Indonesia" from the public Facebook page. \textbf{We converted the video into 6651 images} in a frame-by-frame basis. The colorization results are shown in \textbf{figure \ref{fig:past_photo_best}} where the Proposed 3 models create the most colorful images and clear separation on the object's edges, among other results, followed by ColorNet. On the first image after input, Zhang (2016) et al. tend to create Sephia theme colors (black, white, and brown) while Zhang (2017) et al. tend to be slightly \textit{blue-ish} that happens mostly on all images.
        
        Even though Proposed 3 creates a most compelling color than the rest of the models on landscape images. There is \textbf{a major drawback} especially when the image shows a human object, the proposed model fails to correctly color the human. Proposed 3 creates most of the colors with \textit{blue-ish or green} rather than Sephia theme colors like the others which is more natural as shown in \textbf{figure \ref{fig:past_photo_human_obj}}. However, Proposed 3 often creates more colorful and natural  results on some images with blue color on some parts of image while the rest of the models are more prevalent to be gray or Sephia theme colors (see \textbf{figure \ref{fig:past_photo_best}}). These occur due to the nature of pretrain dataset which mostly contains of scenes, environments, or objects.

\section{Conclusion}
    We have proposed a new architecture for the automatic colorization of grayscale images using a convolutional neural network by combining pretrained U-Net with the Fusion layer. Our fine-tuned U-Net provides the lowest Mean Average Error (MAE) compared to the state-of-the-art models. Moreover, adding a Fusion layer with a classification loss into U-Net can improve the colorization results of plain U-Net with visually more compelling colorization results. We evaluated our model on a large landscape images dataset and showed a slightly better colorization result and the comparable user study evaluation value against the previous models.

\FloatBarrier
\bibliographystyle{ACM-Reference-Format}
\bibliography{my-references}

\end{document}